\documentclass[11pt,a4paper]{article}
\usepackage[hyperref]{acl2021}
\usepackage{times}
\usepackage{latexsym}
\usepackage{booktabs}
\usepackage{multirow}
\usepackage{graphicx}
\usepackage{subcaption}

\usepackage{microtype}

\aclfinalcopy 
\def\aclpaperid{2121} 

\title{Relative Importance in Sentence Processing}

\author{Nora Hollenstein \\
  Center for Language Technology \\
  University of Copenhagen \\
  \texttt{nora.hollenstein@hum.ku.dk} \\\And
  Lisa Beinborn \\
  Computational Linguistics \& Text Mining Lab\\
  Vrije Universiteit Amsterdam \\
  \texttt{l.beinborn@vu.nl} \\}

\date{}

\begin{document}
\maketitle
\begin{abstract}
Determining the relative importance of the elements in a sentence is a key factor for effortless natural language understanding. For human language processing, we can approximate patterns of relative importance by measuring reading fixations using eye-tracking technology. In neural language models, gradient-based saliency methods indicate the relative importance of a token for the target objective. In this work, we compare patterns of relative importance in English language processing by humans and models and analyze the underlying linguistic patterns. We find that human processing patterns in English correlate strongly with saliency-based importance in language models and not with attention-based importance. Our results indicate that saliency could be a cognitively more plausible metric for interpreting neural language models. The code is available on github: \url{https://github.com/beinborn/relative_importance}.
\end{abstract}

\hyphenation{an-tici-pate a-naly-ses}
\section{Introduction}
When children learn to read, they first focus on each word individually and gradually learn to anticipate frequent patterns \cite{blythe2011children}. More experienced readers are able to completely skip words that are predictable from the context and to focus on the more \textit{relevant} words of a sentence \cite{schroeder2015developmental}. Psycholinguistic studies aim at unraveling the characteristics that determine the relevance of a word and find that lexical factors such as word class, word frequency, and word complexity play an important role, but that the effects vary depending on the sentential context \cite{rayner1986lexical}. 

In natural language processing, the relative importance of words is usually interpreted with respect to a specific task. Emotional adjectives are most relevant in sentiment detection \cite{socher2013recursive}, relative frequency of a term is an indicator for information extraction \cite{wu2008tfidf}, the relative position of a token can be used to approximate novelty for summarisation \cite{chopra-etal-2016-abstractive}, and function words play an important role in stylistic analyses such as plagiarism detection \cite{stamatatos2011plagiarism}. Neural language models are trained to be a good basis for any of these tasks and are thus expected to represent a more general notion of relative importance \cite{devlin-etal-2019-bert}. 

Relative importance of the input in neural networks can be modulated by the so-called ``attention" mechanism \cite{bahdanau2014neural}. Analyses of image processing models indicate that attention weights reflect cognitively plausible patterns of visual saliency \cite{xu2015show,coco2012scanpatterns}. Recent research in language processing finds that attention weights are not a good proxy for relative importance because different attention distributions can lead to the same predictions \cite{jain-wallace-2019-attention}. Gradient-based methods such as saliency scores seem to better approximate the relative importance of input words for neural processing models \cite{bastings-filippova-2020-elephant}. 

In this work, we compare patterns of relative importance in human and computational English language processing. We approximate relative importance for humans as the relative fixation duration in eye-tracking data collected in naturalistic language understanding scenarios. In related work, \citet{sood-etal-2020-interpreting} measure the correlation between attention in neural networks trained for a document-level question-answering task and find that the attention in a transformer language model deviates strongly from human fixation patterns. In this work, we instead approximate relative importance in computational models using gradient-based saliency and find that it correlates much better with human patterns. 

\section{Determining Relative Importance} \label{s:importance}
The concept of relative importance of a token for sentence processing encompasses several related psycholinguistic phenomena such as relevance for understanding the sentence, difficulty and novelty of a token within the context, semantic and syntactic surprisal, or domain-specificity of a token. We take a data-driven perspective and approximate the relative importance of a token by the processing effort that can be attributed to it compared to the other tokens in the sentence. 

\subsection{In Human Language Processing}
The sentence processing effort can be approximated indirectly using a range of metrics such as response times in reading comprehension experiments \cite{su2019improving}, processing duration in self-paced reading \cite{linzen2016uncertainty}, and voltage changes in electroencephalography recordings \cite{frank2015erp}. In this work, we approximate relative importance using eye movement recordings during reading because they provide online measurements in a comfortable experimental setup which is more similar to a normal, uncontrolled reading experience. Eye-tracking technology can measure with high accuracy how long a reader fixates each word. The fixation duration and the relative importance of a token for the reader are strongly correlated with reading comprehension \cite{rayner1977visual,malmaud2020bridging}. 

 
\setlength{\belowcaptionskip}{-10pt}
\begin{figure}
    \centering
    \fbox{\includegraphics[width=0.45\textwidth]{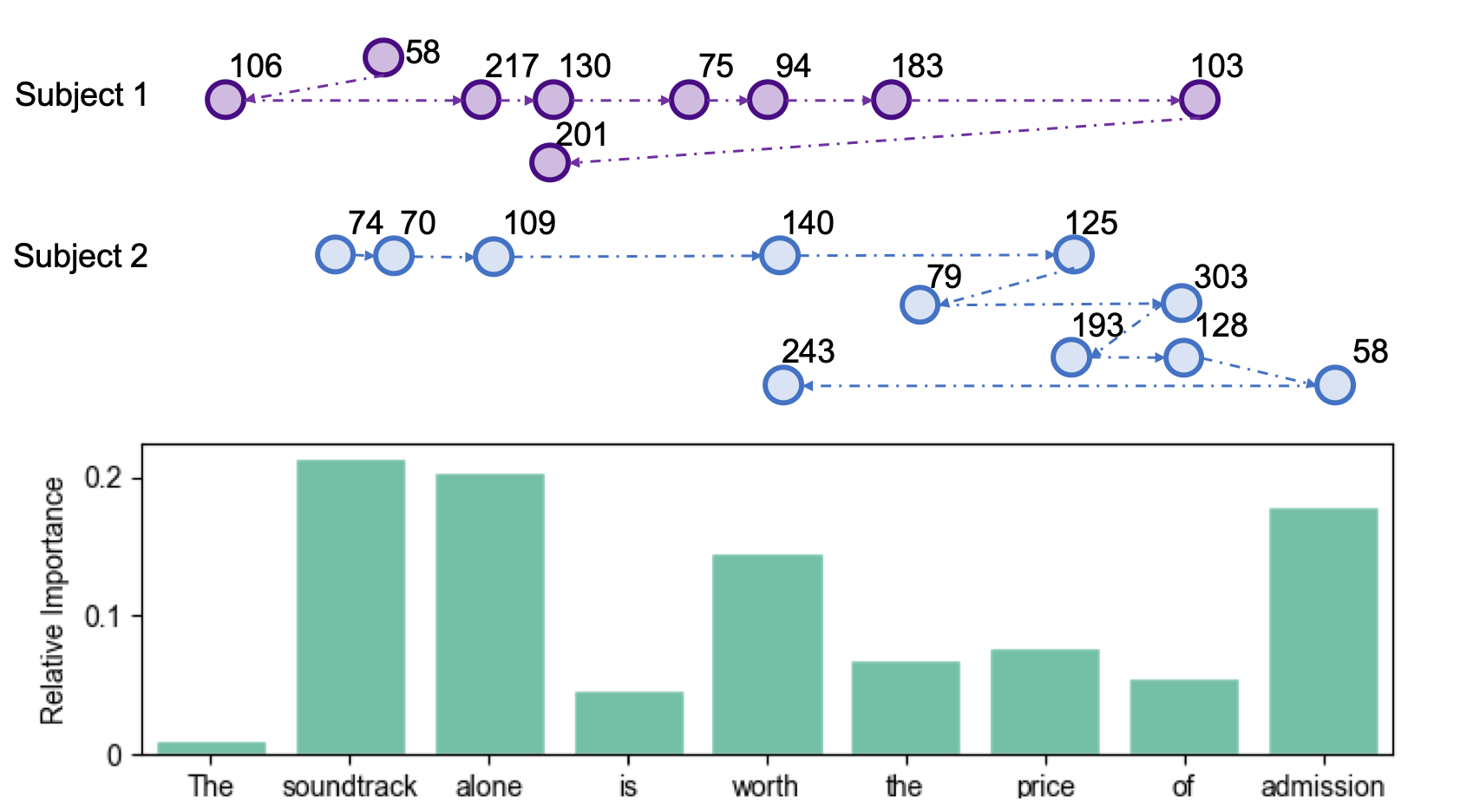}}
    \caption{Example fixations for two subjects in the ZuCo dataset for the sentence ``The soundtrack alone is worth the price of admission". The numbers indicate the fixation duration and the circles represent the approximate horizontal position of the fixation (positions are simplified for better visualization). The plot at the bottom indicates the relative importance of each token averaged over all subjects. }
    \label{fig:fixation}
\end{figure}

Language models that look ahead and take both the left and right context into account are often considered cognitively less plausible because humans process language incrementally from left to right \cite{merkx2020comparing}. However, in human reading, we frequently find regressions: humans fixate relevant parts of the left context again while already knowing what comes next \cite{rayner1998eye}. In Figure \ref{fig:fixation}, subject 1 first reads the entire sentences and then jumps back to the token ``alone". Subject 2 performs several regressions to better understand the second half of the sentence. The fixation duration is a cumulative measure that sums over these repeated fixations. Absolute fixation duration can vary strongly between subjects due to differences in reading speed but the relative fixation duration provides a good approximation for the relative importance of a token as it abstracts from individual differences. We average the relative fixation duration over all subjects to obtain a more robust signal (visualized in the plot at the bottom of Figure \ref{fig:fixation}). 


\subsection{In Computational Language Processing}
In computational language models, the interpretation of a token depends on the tokens in its context but not all tokens are equally important. To account for varying importance, so-called attention weights regulate the information flow in neural networks \cite{bahdanau2014neural}. These weights are optimized with respect to a target objective and higher attention for an input token has been interpreted as higher importance with respect to the output \cite{vig-2019-multiscale}. Recent research indicates that complementary attention distributions can lead to the same model prediction \cite{jain-wallace-2019-attention,wiegreffe-pinter-2019-attention} and that the removal of input tokens with large attention weights often does not lead to a change in the model's prediction \cite{serrano-smith-2019-attention}. In transformer models, the attention weights often approximate an almost uniform distribution in higher model layers \cite{abnar-zuidema-2020-quantifying}. \citet{bastings-filippova-2020-elephant} argue that saliency methods are more suitable for assigning importance weights to input tokens. 

Saliency methods calculate the gradient of the output corresponding to the correct prediction with respect to an input element to identify those parts of the input that have the biggest influence on the prediction \cite{lipton2018interpretability}. Saliency maps were first developed for image processing models to highlight the areas of the image that are discriminative with respect to the tested output class \cite{simonyan2014saliency}. \citet{li-2016-erasure} adapt this method to calculate the relative change of the output probabilities with respect to individual input tokens in text classification tasks and \citet{ding-etal-2019-saliency} calculate saliency maps for interpreting the alignment process in machine translation models. 

In general-purpose language models such as BERT \cite{devlin-etal-2019-bert}, the objective function tries to predict a token based on its context. A saliency vector for a masked token thus indicates the importance of each of the tokens in the context of correctly predicting the masked token \cite{madsen2019visualizing}. 

We iterate over each token vector $\mathbf{x}_i$ in our input sequence $x_1$, $x_2$, ... $x_n$. Let $\mathbf{X}_i$ be the input matrix with $\mathbf{x}_i$ being masked. 
The saliency $s_{ij}$ for input token $\mathbf{x}_j$ for the prediction of the correct token $\mathbf{t}_i$ is then calculated as the Euclidean norm of the gradient of the logit for $x_i$. 
\begin{equation}\label{eq:grad}
    s_{ij} = \|\nabla_{\mathbf{x}_j} f_{t_i}(\mathbf{X}_i)\|_2
\end{equation}

The saliency vector $\mathbf{s}_i$ indicates the relevance of each token for the correct prediction of the masked token $t_i$.\footnote{Our implementation adapts code from  \url{https://pypi.org/project/textualheatmap/}. An alternative would be to multiply saliency and input \cite{alammar2020explaining}.} The saliency scores are normalized by dividing by the maximum. We determine the relative importance of a token by summing over the saliency scores for each token. For comparison, we also approximate importance using attention values from the last layer of each model as \citet{sood-etal-2020-interpreting}.   

 

\subsection{Patterns of Relative Importance}
Relative importance in human processing and in computational models is sensitive to linguistic properties. \citet{rayner1998eye} provides a detailed overview of token-level features that have been found to correlate with fixation duration such as length, frequency, and word class. On the contextual level, lexical and syntactic disambiguation processes cause regressions and thus lead to longer fixation duration \cite{just1980theory, lowder2018lexical}. Computational models are also highly susceptible to frequency effects and surprisal metrics calculated using language models can predict the human processing effort \cite{frank2013word}.

The inductive bias of language processing models can be improved using the eye-tracking signal \cite{barrett-etal-2018-sequence,klerke-plank-2019-glance} and the modification leads to more ``human-like" output in generative tasks \cite{takmaz-etal-2020-generating,sood-2020-neurips}. This indicates that patterns of relative importance in computational representations differ from human processing patterns. 
Previous work focused on identifying links between the eye-tracking signal and attention \cite{sood-etal-2020-interpreting}. To our knowledge, this is the first attempt to correlate fixation duration with saliency metrics.

The eye-tracking signal represents human reading processes aimed at language understanding. In previous work, we have shown that contextualized language models can predict eye patterns associated with human reading \cite{hollenstein-etal-2021-multilingual}, which indicates that computational models and humans encode similar linguistic patterns. It remains an open debate to which extent language models are able to approximate language understanding \cite{bender-koller-2020-climbing}. We are convinced that language needs to be cooperatively grounded in the real world \cite{beinborn-etal-2018-multimodal}. Purely text-based language models clearly miss important aspects of language understanding but they can approximate human performance in an impressive range of processing tasks. We aim to gain a deeper understanding of the similarities and differences between human and computational language processing to better evaluate the capabilities of language models.

 \setlength{\tabcolsep}{5pt}
\begin{table}
\small
\centering
\begin{tabular}{llrrrcr}
\toprule
 &Dataset& BERT & Distil & ALBERT  & Rand \\
\midrule
\multirow{2}{*}{\textbf{Saliency}} &GECO & \textbf{.54} & .51 & .48  &.00 \\
&ZuCo & \textbf{.68} & .64 & .62 & .00\\
\midrule
\multirow{2}{*}{\textbf{Attention}} &GECO & .18 &.06&.26&.00 \\
& ZuCo & .11&.03&.37&.00 \\
\bottomrule
\end{tabular}
\caption{Spearman correlation between relative fixation duration by humans and attention and saliency in the language models. Correlation values are averaged over all sentences. \textit{Rand} is a permutation baseline. }
\label{t:correlation_results}
\end{table}
\setlength{\tabcolsep}{7pt}

\section{Methodology}
We extract relative importance values for tokens from eye-tracking corpora and language models as described in section \ref{s:importance} and calculate the Spearman correlation for each sentence.\footnote{Kendall's $\tau$ and KL divergence yield similar results. } We first average the correlation over all sentences to analyze whether the importance patterns of humans and models are comparable and then conduct token-level analyses. 
\subsection{Eye-tracking Corpora}
We extract the relative fixation duration from two eye-tracking corpora and average it over all readers for each sentence. Both corpora record natural reading and the text passages were followed by multiple-choice questions to test the readers' comprehension.

\paragraph{GECO} contains eye-tracking data from 14 native English speakers reading the entire novel \textit{The Mysterious Affair at Styles} by Agatha Christie \cite{cop2017presenting}. The text was presented on the screen in paragraphs. 

\paragraph{ZuCo} contains eye-tracking data of 30 native English speakers reading full sentences from movie reviews and Wikipedia articles \cite{hollenstein2018zuco,hollenstein-etal-2020-zuco}.\footnote{We combine ZuCo 1.0 (T1, T2) and  ZuCo 2.0. (T1).} 

\subsection{Language Models}
We compare three state-of-the-art language models trained for English: BERT, ALBERT, and DistilBERT.\footnote{We use the \textit{Huggingface} transformers implementation \cite{wolf-etal-2020-transformers} and the models \texttt{bert-based-uncased}, \texttt{albert-base-v2}, and \texttt{distilbert-base-uncased}.} BERT was the first widely successful transformer-based language model and remains highly influential \cite{devlin-etal-2019-bert}. ALBERT and DistilBERT are variants of BERT that require less training time due to a considerable reduction of the training parameters while maintaining similar performance on benchmark datasets \cite{lan2019albert,sanh2019distilbert}.\footnote{Reduction is achieved by parameter sharing across layers (ALBERT) and by distillation which approximates the output distribution of the original BERT model using a smaller network (DistilBERT). See model references for details.} 
We analyze if the lighter architectures have an influence on the patterns of relative importance that the models learn.  

\begin{figure*}
\centering
\begin{subfigure}{0.49\textwidth}
\centering
\includegraphics[width=0.9\linewidth]{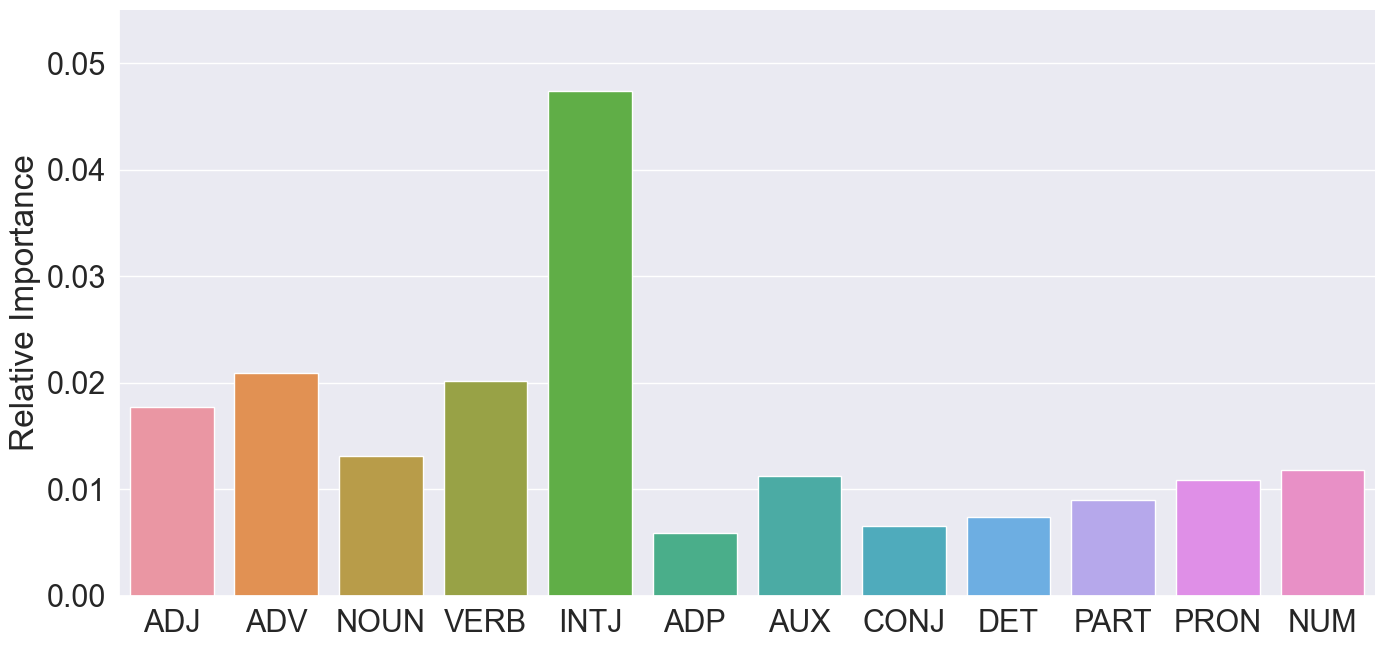} 
\caption{Human Fixation}
\end{subfigure}
\begin{subfigure}{0.49\textwidth}
\centering
\includegraphics[width=0.9\linewidth]{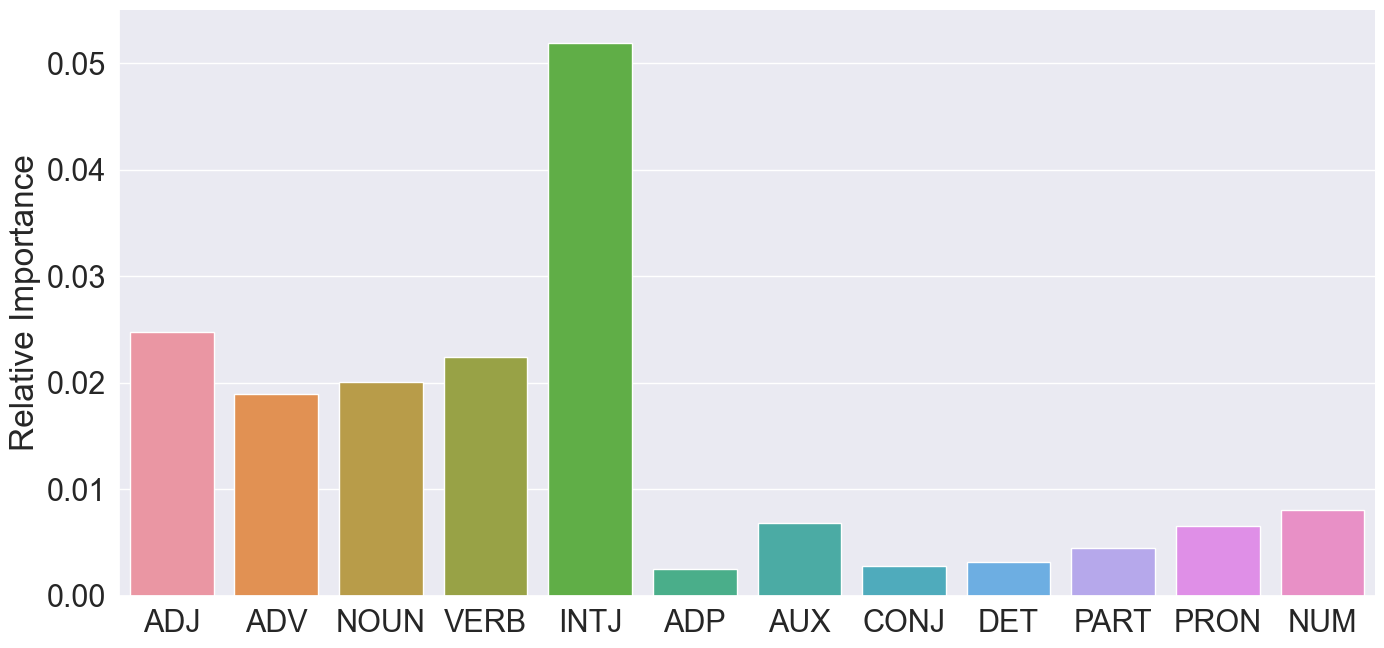}
\caption{Model saliency}
\end{subfigure}
\caption{Relative importance of tokens with respect to word class. Relative importance is measured as relative fixation duration for humans in GECO (left) and as relative gradient-based saliency in the BERT model (right). 
}
\label{fig:wordclasses}
\end{figure*}


\section{Results}
The results in Table \ref{t:correlation_results} show that relative fixation duration by humans strongly correlates with the saliency values of the models. In contrast, attention-based importance does not seem to be able to capture the human importance pattern. A random permutation baseline that shuffles the importance assigned by the language model yields no correlation (0.0) in all conditions.\footnote{We repeat the permutation 100 times and average the correlation over all iterations.} As the standard deviations of the correlation across sentences are quite high (ZuCo: $\sim$0.22, GECO: $\sim$0.39), the small differences between models can be neglected (although they are consistent across corpora). For the subsequent analyses, we focus only on the BERT model which yields the best results. The differences between the corpora might be related to the number of sentences and the differences in average sentence length (ZuCo: 924, 19.5, GECO: 4,926, 12.7). 

\paragraph{Length and Frequency} 
In eye-tracking data, word length correlates with fixation duration because it takes longer to read all characters. The correlation for frequency is inverse because high-frequency words (e.g. “the”, “has”) are often skipped in processing as they carry (almost) no meaning \cite{rayner1998eye}. For English, word frequency and word length are both closely related to word complexity \cite{beinborn-etal-2014-predicting}. 
Language models do not directly encode word length but they are sensitive to word frequency.

Our results in Table \ref{t:lenfreq} show that both token length and frequency are strongly correlated with relative importance on the sentence level. Interestingly, the correlation decreases when it is calculated directly over all tokens indicating that the token-level relation between length and importance is more complex than the correlation might suggest. 


\begin{table}
\small
\centering
\begin{tabular}{llrrrr}
\toprule
& &\multicolumn{2}{c}{\textbf{Length}}& \multicolumn{2}{c}{\textbf{Frequency}}\\
 & &Sent & Tok & Sent & Tok\\
 \midrule
\multirow{2}{*}{GECO} & Human & .69  & .31& -.36&-.25\\
& BERT & .65  & .27& -.48 & -.28\\
\midrule
\multirow{2}{*}{ZuCO} &Human & .75 &.47& -.52&-.36\\
& BERT & .72  &.36 & -.65& -.40\\
\bottomrule
\end{tabular}
\caption{Spearman correlation between relative importance and word length and frequency. For the \textit{Sent} condition, correlation is calculated per sentence and averaged. For \textit{Tok}, importance is normalized by sentence length and correlation is calculated over all tokens.}
\label{t:lenfreq}
\end{table}

\begin{figure}
    \centering
    \includegraphics[trim=15 5 5 5,width=0.46\textwidth]{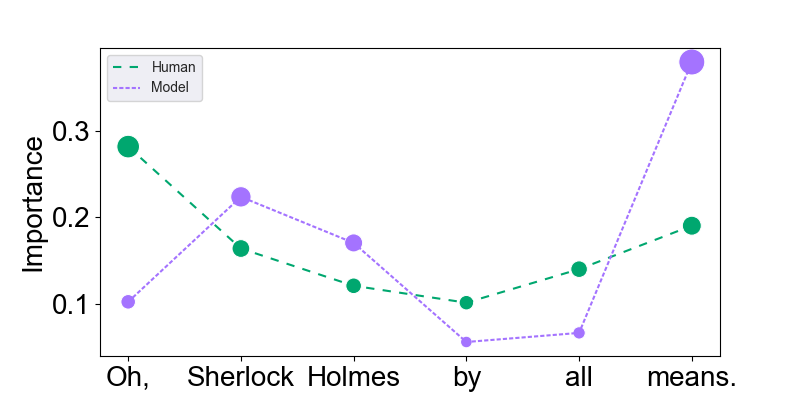}
    \caption{Relative importance values for an example sentence from the GECO corpus for the BERT model and the human averages.  
    }
    \label{fig:example}
\end{figure}

\paragraph{Word Class}
Figure \ref{fig:wordclasses} shows the average relative importance of all tokens belonging to the same word class (normalized by sentence length). We see that both humans and BERT clearly assign higher importance to content words (left) than to function words (right). Interjections such as ``Oh" in figure \ref{fig:example} receive the highest relevance which is understandable because they interrupt the reading flow. When we look at individual sentences, we note that the differences in importance are more pronounced in the model saliency while human fixation duration yields a smoother distribution over the tokens.    

\paragraph{Novelty}
We extract the language model representations for each sentence separately whereas the readers processed the sentences consecutively. If tokens are mentioned repeatedly such as ``Sherlock Holmes" which also occurred in the sentence preceding the example in Figure \ref{fig:example}), processing ease increases for the reader, and not for the model. Some language models are able to process multiple sentences, but establishing semantic links across sentences remains a challenge. 

\section{Conclusion}
We find that human sentence processing patterns in English correlate strongly with saliency-based importance in language models and not with attention-based importance. Our results indicate that saliency could be a cognitively more plausible metric for interpreting neural language models. In future work, it would be interesting to test the robustness of the approach with different variants for calculating saliency \cite{bastings-filippova-2020-elephant,ding-koehn-2021-evaluating}. 
As we conducted our analyses only for English data, it is not yet clear whether our results generalize across languages. We will address this in future work using eye-tracking data from non-English readers \cite{makowski2018discriminative,laurinavichyute2019russian} and comparing mono- and multilingual models \cite{beinborn-choenni-2020-semantic}. We want to extend the token-level analyses to syntactic phenomena and cross-sentence effects. For example, it would be interesting to see how a language model encodes relative importance for sentences that are syntactically correct but not semantically meaningful \cite{gulordava-etal-2018-colorless}. 

Previous work has shown that the inductive bias of recurrent neural networks can be modified to obtain cognitively more plausible model decisions \cite{bhatt-etal-2020-much,shen2018ordered}. In principle, our approach can also be applied to left-to-right models such as GPT-2 \cite{radford2019language}. In this case, the tokens at the beginning of the sentence would be assigned disproportionately high importance as the following tokens cannot contribute to the prediction of preceding tokens in incremental processing. It might thus be more useful to only use the first fixation duration of the gaze signal for analyzing importance in left-to-right models. However, we think that the regressions by the readers provide valuable information about sentence processing.

\section{Ethical Considerations}
Data from human participants were leveraged from freely available datasets \citep{hollenstein2018zuco,hollenstein-etal-2020-zuco,cop2017presenting}. The datasets provide anonymized records in compliance with ethical board approvals and do not contain any information that can be linked to the participants.

\section*{Acknowledgements}
Lisa Beinborn's research was partially funded by the Dutch National Science Organisation (NWO) through the project CLARIAH-PLUS (CP-W6-19-005). 
We thank the anonymous reviewers for their constructive feedback. 
\bibliography{anthology,clap}
\bibliographystyle{acl_natbib}

\onecolumn
\appendix
\section{Additional Results}
\label{sec:additional-results}

\begin{figure*}[h]
\centering
\begin{subfigure}{0.49\textwidth}
\includegraphics[width=0.95\linewidth]{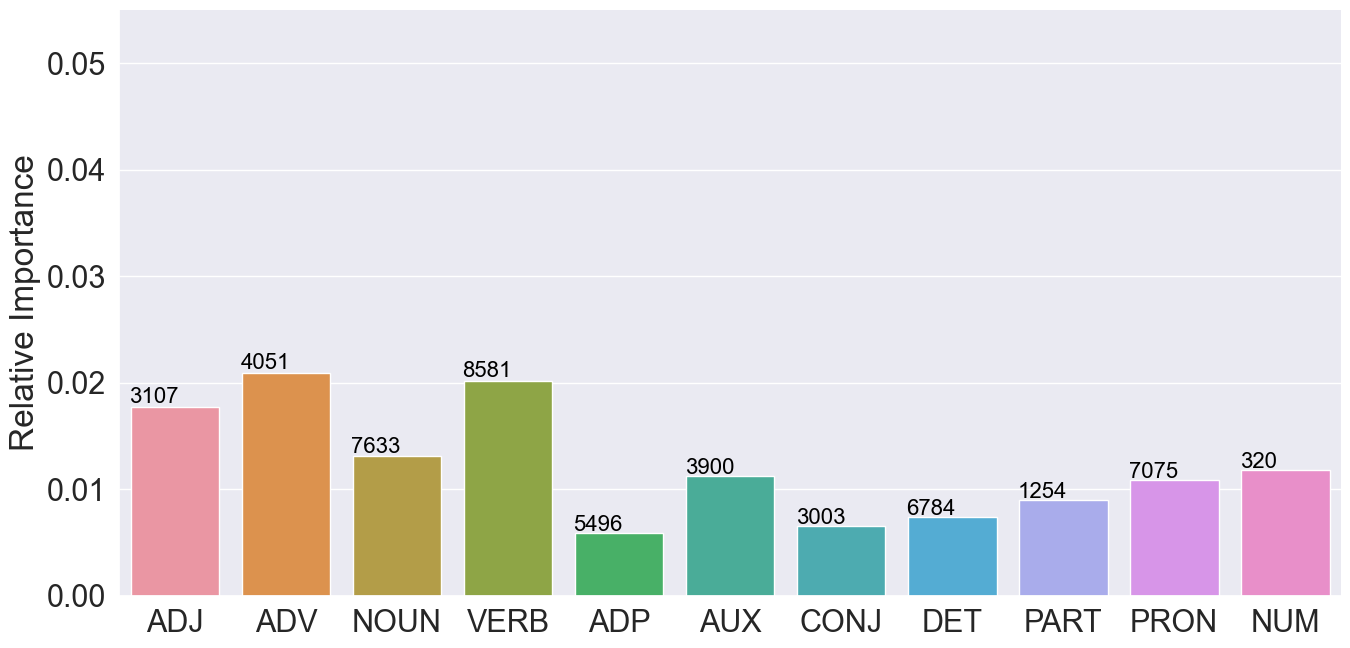} 
\caption{Human Fixation}
\end{subfigure}
\begin{subfigure}{0.49\textwidth}
\includegraphics[width=0.95\linewidth]{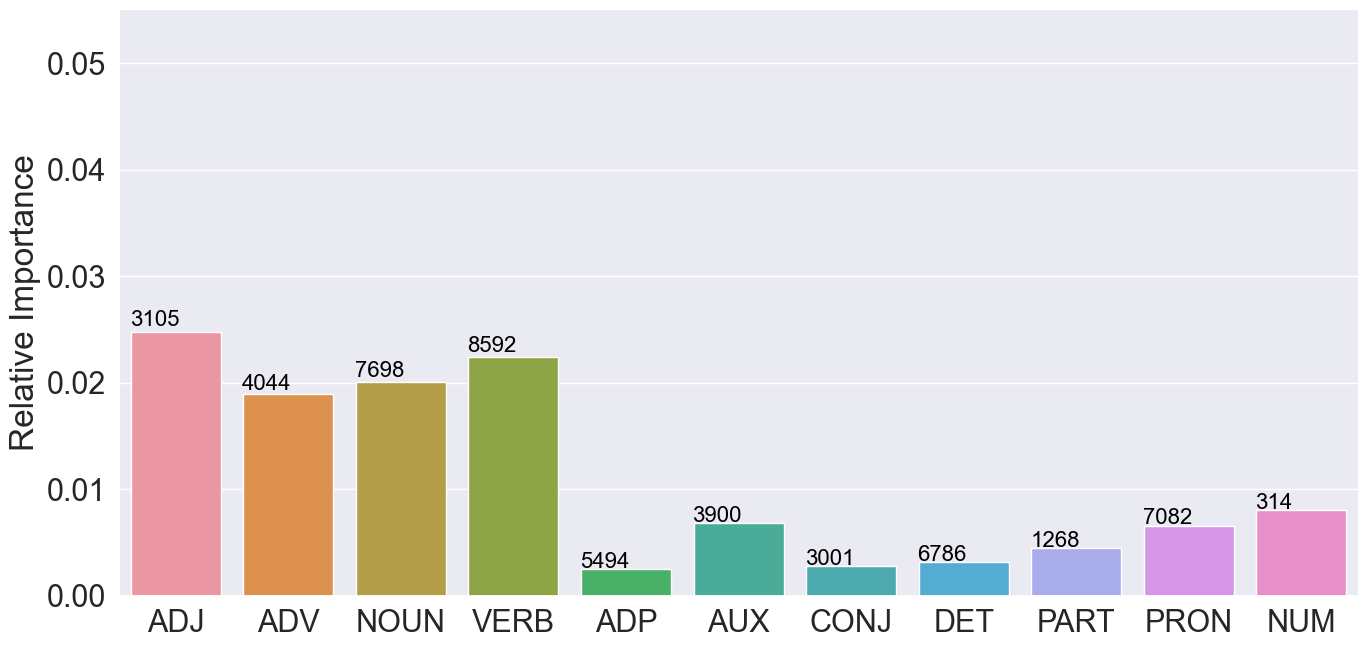}
\caption{Model saliency}
\end{subfigure}
\caption{Relative importance of tokens with respect to word class in the GECO dataset. Relative importance is measured as relative fixation duration for humans (top) and as relative gradient-based saliency in the BERT model (bottom). This is the same figure as Figure \ref{fig:wordclasses} in the paper but it includes the number of instances per word class on top of the respective bar. }
\label{fig:wordclassesgeco}
\end{figure*}

\vspace{2cm}

\begin{figure*}[h]
    \centering
    \includegraphics[width=0.77\textwidth]{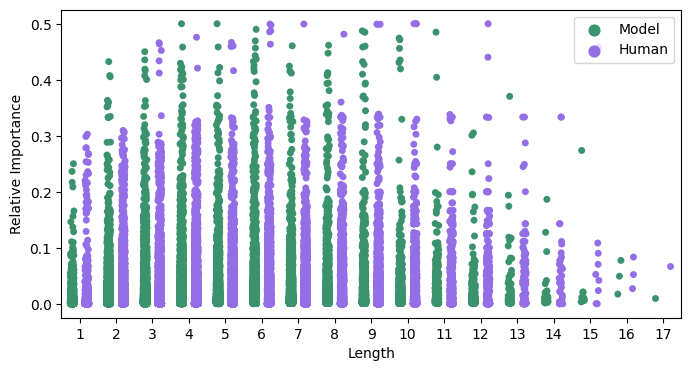}
    \caption{Relative importance values with respect to word length from human readers and from the BERT model for the GECO corpus.}
    \label{fig:length}
\end{figure*}

\newpage
\begin{figure*}[h]
    \centering
    \includegraphics[width=0.7\textwidth]{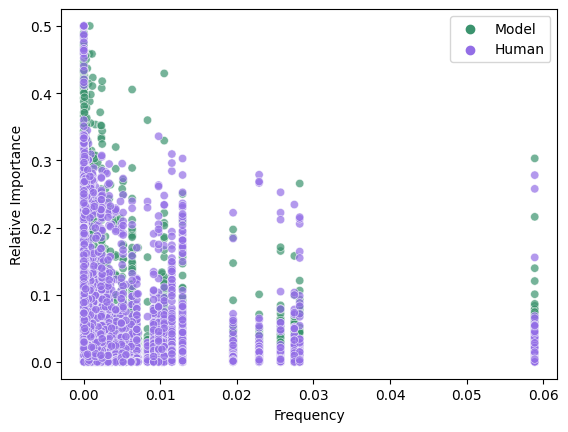}
    \caption{Relative importance values with respect to word frequency from human readers and from the BERT model for the GECO corpus.}
    \label{fig:frequency}
\end{figure*}

\end{document}